\documentclass[11pt,a4paper]{article}
\usepackage[hyperref]{acl2018}
\usepackage{times}
\usepackage{latexsym}
\usepackage{algorithm}
\usepackage[noend]{algpseudocode}
\usepackage{xcolor,colortbl}

\usepackage[utf8]{inputenc} % allow utf-8 input
\usepackage[T1]{fontenc}    % use 8-bit T1 fonts
\usepackage{url}            % simple URL typesetting
\usepackage{booktabs}       % professional-quality tables
\usepackage{amsfonts}       % blackboard math symbols
\usepackage{nicefrac}       % compact symbols for 1/2, etc.
\usepackage{microtype}      % microtypography
\usepackage{amsmath}
\usepackage{graphicx}
\usepackage{tabularx}
\usepackage{subcaption}

\aclfinalcopy % Uncomment this line for the final submission
\def\aclpaperid{} %  Enter the acl Paper ID here

\title{On the Limitations of Unsupervised Bilingual Dictionary Induction}
\author  
  {
	\begin{tabular}{ccc}
	Anders Søgaard$^{\heartsuit}$ & Sebastian Ruder$^{\spadesuit\clubsuit}$ & Ivan Vulić$^{\Diamond}$\\
	\end{tabular}
	\\
    $^\heartsuit$University of Copenhagen, Copenhagen, Denmark\\
    $^\spadesuit$Insight Research Centre, National University of Ireland, Galway, Ireland\\
    $^\clubsuit$Aylien Ltd., Dublin, Ireland\\
    $^\Diamond$Language Technology Lab, University of Cambridge, UK\\
	{\tt \small{soegaard@di.ku.dk,sebastian@ruder.io,iv250@cam.ac.uk}}
}

\date{}

\begin{document}
\maketitle
\begin{abstract}
%{\bf Alternative abstract:} 
Unsupervised machine translation---i.e., not assuming {\em any} cross-lingual supervision signal, whether a dictionary, translations, or comparable corpora---seems impossible, but nevertheless, \newcite{Lample2017} recently proposed a fully unsupervised machine translation (MT) model. The model relies heavily on an adversarial, unsupervised alignment of word embedding spaces for {\em bilingual dictionary induction} \cite{Lample2018crosslingual}, which we examine here.
% We determine the circumstances under which the technique performs well, and when it does not.
Our results identify the limitations of current unsupervised MT: unsupervised bilingual dictionary induction performs much worse on morphologically rich languages that are not dependent marking, when monolingual corpora from different domains or different embedding algorithms are used. We show that a simple trick, exploiting a weak supervision signal from identical words, enables more robust induction, and establish a near-perfect correlation between unsupervised bilingual dictionary induction performance and a previously unexplored graph similarity metric. % that has not been used to study word embeddings before. 

\end{abstract}

\section{Introduction}

Cross-lingual word representations enable us to reason about word meaning in multilingual contexts and facilitate cross-lingual transfer \cite{Ruder2018survey}. Early cross-lingual word embedding models relied on large amounts of parallel data \cite{Klementiev,Mikolov2013d}, but more recent approaches have tried to minimize the amount of supervision necessary \cite{IvanSeed,Levy:ea:17,Artetxe2017}. Some researchers have even presented {\em unsupervised}~methods that do not rely on any form of cross-lingual supervision at all \cite{Barone2016,Lample2018crosslingual,Zhang2017c}.

Unsupervised cross-lingual word embeddings hold promise to induce bilingual lexicons and machine translation models in the absence of dictionaries and translations \cite{Barone2016,Zhang2017c,Lample2017}, and would therefore be a major step toward machine translation to, from, or even between low-resource languages. 

Unsupervised approaches to learning cross-lingual word embeddings are based on the assumption that monolingual word embedding graphs are approximately isomorphic, that is, after removing a small set of vertices (words) \cite{Mikolov2013,Barone2016,Zhang2017c,Lample2018crosslingual}. In the words of \newcite{Barone2016}:

\vspace{1mm}
\begin{center}
{\small
\begin{minipage}{2.5in}{\em \ldots we hypothesize that, if languages
are used to convey thematically similar information
in similar contexts, these random processes
should be approximately isomorphic between
languages, and that this isomorphism can
be learned from the statistics of the realizations of
these processes, the monolingual corpora, in principle
without any form of explicit alignment.}\end{minipage}}%
\end{center}

\noindent Our results indicate this assumption is not true in general, and that approaches based on this assumption have important limitations. 

\paragraph{Contributions} We focus on the recent state-of-the-art unsupervised model of \newcite{Lample2018crosslingual}.\footnote{Our motivation for this is that \newcite{Artetxe2017} use small dictionary seeds for supervision, and \newcite{Barone2016} seems to obtain worse performance than \newcite{Lample2018crosslingual}. Our results should extend to \newcite{Barone2016} and \newcite{Zhang2017c}, which rely on very similar methodology.} Our contributions are: (a) In \S\ref{sec:how_similar}, we show that the monolingual word embeddings used in \newcite{Lample2018crosslingual} are {\em not}~approximately isomorphic, using the VF2 algorithm \cite{Cordella:ea:01} and we therefore introduce a metric for quantifying the similarity of word embeddings, based on Laplacian eigenvalues. (b) In \S\ref{sec:assumptions}, we identify circumstances under which the unsupervised bilingual dictionary induction (BDI) algorithm proposed in \newcite{Lample2018crosslingual} does not lead to good performance. (c) We show that a simple trick, exploiting a weak supervision signal from words that are identical across languages, makes the algorithm much more robust. Our main finding is that the performance of unsupervised BDI depends heavily on all three factors: the language pair, the comparability of the monolingual corpora, and the parameters of the word embedding algorithms. 

%We show that monolingual embeddings In the following, we will propose probes that target specific aspects of these zero-supervision methods in order to reveal the assumptions that they implicitly make. In particular, these methods require that the monolingual embeddings of different languages have the same distribution and can be mapped onto each other.

\section{How similar are embeddings across languages?} \label{sec:how_similar}

As mentioned, recent work focused on unsupervised BDI assumes that monolingual word embedding spaces (or at least the subgraphs formed by the most frequent words) are approximately isomorphic. In this section, we show, by investigating the nearest neighbor graphs of word embedding spaces, that word embeddings are far from isomorphic. We therefore introduce a method for computing the similarity of non-isomorphic graphs. In \S\ref{sec:evaluation_eigenvector}, we correlate our similarity metric with performance on unsupervised BDI. 	%, and we discuss the idea of frequent outliers servicng as anchors. 

\begin{figure}[t!]
    \centering
    \begin{subfigure}[t]{0.19\textwidth}
        \centering
       \includegraphics[height=1.2in]{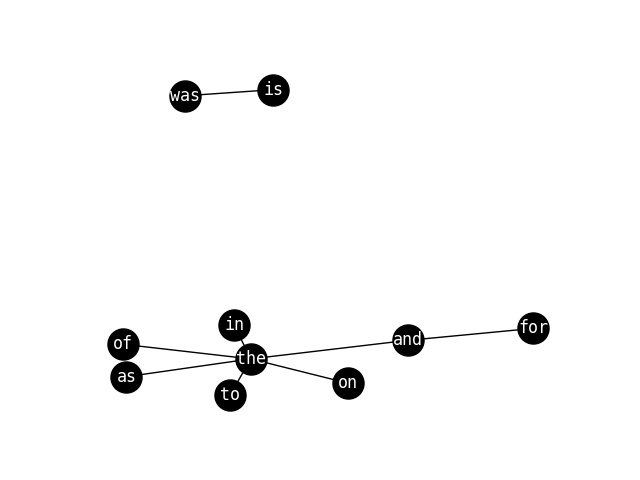}
        %\caption{German graph} \label{fig:word_par}
        \caption{Top 10 most frequent English words}
    \end{subfigure}%
    ~ 
    \begin{subfigure}[t]{0.19\textwidth}
        \centering
       \includegraphics[height=1.2in]{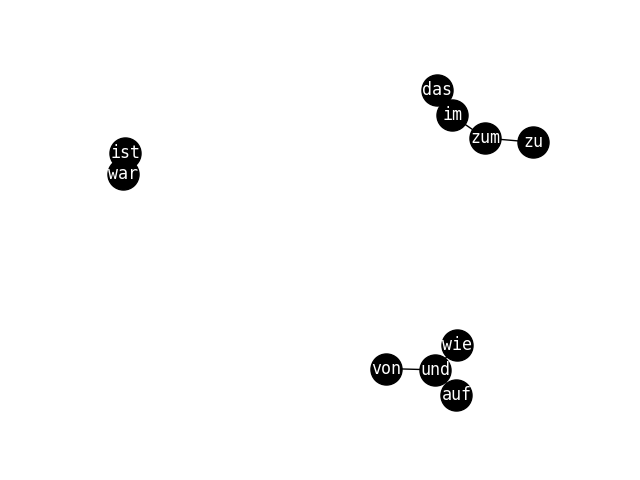}
        %\caption{English graph} \label{fig:word_comp}
        \caption{German translations of top 10 most frequent English words}
    \end{subfigure}
        \begin{subfigure}[t]{0.19\textwidth}
        \centering
       \includegraphics[height=1.2in]{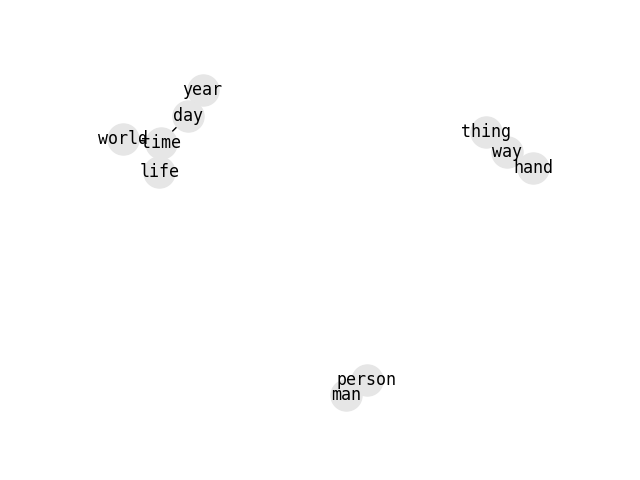}
        %\caption{German graph} \label{fig:word_par}
        \caption{Top 10 most frequent English nouns}
    \end{subfigure}%
    ~ 
    \begin{subfigure}[t]{0.19\textwidth}
        \centering
       \includegraphics[height=1.2in]{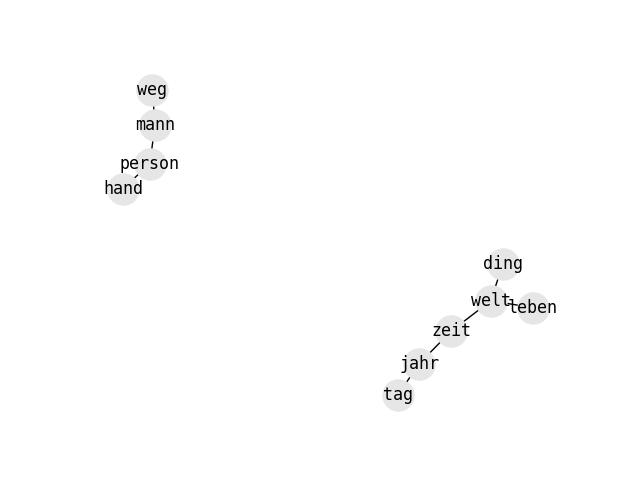}
        %\caption{English graph} \label{fig:word_comp}
         \caption{German translations of top 10 most frequent English nouns}
    \end{subfigure}
    \caption{\label{plots}Nearest neighbor graphs.} \label{fig:nn-graphs}
    \vspace{-3.0mm}
\end{figure}

\paragraph{Isomorphism} To motivate our study, we first establish that word embeddings are far from graph isomorphic\footnote{Two graphs that contain the same number of graph vertices connected in the same way are said to be isomorphic. In the context of weighted graphs such as word embeddings, a weak version of this is to require that the underlying nearest neighbor graphs for the most frequent $k$ words are isomorphic.}---even for two closely related languages, English and German, and using embeddings induced from comparable corpora (Wikipedia) with the same hyper-parameters. 

If we take the top $k$ most frequent words in English, and the top $k$ most frequent words in German, and build nearest neighbor graphs for English and German using the monolingual word embeddings used in \newcite{Lample2018crosslingual}, the graphs are of course very different. This is, among other things, due to German case and the fact that {\em the} translates into {\em der}, {\em die}, and {\em das}, but unsupervised alignment does not have access to this kind of information. {\em Even}~if we consider the top $k$ most frequent English words {\em and their translations}~into German, the nearest neighbor graphs are not isomorphic. Figure~\ref{plots}a-b shows the nearest neighbor graphs of the top 10 most frequent English words on Wikipedia, and their German translations. 

Word embeddings are particularly good at capturing relations between nouns, but even if we consider the top $k$ most frequent English {\em nouns}~and their translations, the graphs are not isomorphic; see Figure~\ref{plots}c-d. We take this as evidence that word embeddings are not approximately isomorphic across languages. We also ran graph isomorphism checks on 10 random samples of frequent English nouns and their translations into Spanish, and only in 1/10 of the samples were the corresponding nearest neighbor graphs isomorphic. 

\paragraph{Eigenvector similarity} Since the nearest neighbor graphs are not isomorphic, even for frequent translation pairs in neighboring languages, we want to quantify the potential for unsupervised BDI using a metric that captures varying degrees of graph similarity. Eigenvalues are compact representations of global properties of graphs, and we introduce a spectral metric based on Laplacian eigenvalues \cite{Shigehalli2011} that quantifies the extent to which the nearest neighbor graphs are {\em isospectral}. Note that (approximately) isospectral graphs need not be (approximately) isomorphic, but (approximately) isomorphic graphs are always (approximately) isospectral \cite{Gordon:ea:92}. Let $A_1$ and $A_2$ be the adjacency matrices of the nearest neighbor graphs $G_1$ and $G_2$ of our two word embeddings, respectively. Let $L_1 = D_1 - A_1$ and $L_2 = D_2 - A_2$ be the Laplacians of the nearest neighbor graphs, where $D_1$ and $D_2$ are the corresponding diagonal matrices of degrees. We now compute the eigensimilarity of the Laplacians of the nearest neighbor graphs, $L_1$ and $L_2$. For each graph, we find the smallest $k$ such that the sum of the $k$ largest Laplacian eigenvalues is <$90$\% of the Laplacian eigenvalues. We take the smallest $k$ of the two, and use the sum of the squared differences between the largest $k$ Laplacian eigenvalues $\Delta$ as our similarity metric. 

$$\Delta = \sum^k_{i=1} (\lambda_{1_i} - \lambda_{2_i})^2$$

where $k$ is chosen s.t.

$$\min_j \{ \frac{\sum^k_{i=1} \lambda_{j_i}}{\sum^n_{i=1}
\lambda_{ji}} > 0.9\}$$

Note that $\Delta=0$ means the graphs are isospectral, and the metric goes to infinite. Thus, the higher $\Delta$ is, the {\em less}~similar the graphs (i.e., their Laplacian spectra). We discuss the correlation between unsupervised BDI performance and approximate isospectrality or eigenvector similarity in \S\ref{sec:evaluation_eigenvector}. 

%\paragraph{Anchors} We have established that word embeddings are far from isomorphic. This means any mapping or projection will be noisy and approximate; something we hope to capture better using eigenvector similarity. But this also means that the signal for learning this projection must be either the {\em global geometric properties of the embeddings} or {\em a set of highly frequent outlier words} that drive the projection. Examples of such outlier words could be punctuation, dummy symbols used for generalization in tokenization, or very frequent loanwords such as {\em Wikipedia} or {\em www}.\footnote{We leave out the hypothesis that unsupervised bilingual dictionary induction works because of numerals, as in \newcite{Artetxe2017}, because numerals are replaced by a dummy symbol in the FastText embeddings \cite{Bojanowski2017}.} We investigate whether unsupervised bilingual dictionary induction is sensitive to anchors in \S\ref{sec:anchors}. 

\section{Unsupervised cross-lingual learning} \label{sec:assumptions}

\subsection{Learning scenarios}

Unsupervised neural machine translation relies on BDI using cross-lingual embeddings \cite{Lample2017,ArtetxeNMT}, which in turn relies on the assumption that word embedding graphs are approximately isomorphic. The work of \newcite{Lample2018crosslingual}, which we focus on here, also makes several implicit assumptions that may or may not be necessary to achieve such isomorphism, and which may or may not scale to low-resource languages. The algorithms are not intended to be limited to learning scenarios where these assumptions hold, but since they do in the reported experiments, it is important to see to what extent these assumptions are necessary for the algorithms to produce useful embeddings or dictionaries. 

We focus on the work of \newcite{Lample2018crosslingual}, who present a fully unsupervised approach to aligning monolingual word embeddings, induced using {\em fastText}~\cite{Bojanowski2017}. We describe the learning algorithm in \S\ref{sec:conneau_approach}. \newcite{Lample2018crosslingual} consider a specific set of learning scenarios:

%\ldots
\begin{itemize}
\item[(a)] The authors work with the following {\bf languages}: English-\{French, German, Chinese, Russian, Spanish\}. These languages, except French, are dependent marking (Table~\ref{langs}).\footnote{A dependent-marking language marks agreement and case more commonly on dependents than on heads.} We evaluate \newcite{Lample2018crosslingual} on (English to) Estonian ({\sc et}), Finnish ({\sc fi}), Greek ({\sc el}), Hungarian ({\sc hu}), Polish ({\sc pl}), and Turkish ({\sc tr}) in \S\ref{sec:impact_lang_sim}, to test whether the selection of languages in the original study introduces a bias.
\item[(b)] The monolingual corpora in their experiments are comparable; Wikipedia corpora are used, except for an experiment in which they include Google Gigawords. We evaluate across different {\bf domains}, i.e., on all combinations of Wikipedia, EuroParl, and the EMEA medical corpus, in \S\ref{sec:impact_domain_diff}. We believe such scenarios are more realistic for low-resource languages. 
\item[(c)] The monolingual embedding models are induced using the same \textbf{algorithms} with the same {\bf hyper-parameters}. We evaluate \newcite{Lample2018crosslingual} on pairs of embeddings induced with different hyper-parameters in \S\ref{sec:impact_hyperparam}. While keeping hyper-parameters fixed is always possible, it is of practical interest to know whether the unsupervised methods work on any set of pre-trained word embeddings. 
\end{itemize}
%To motivate our line of interrogation and show that such an assumption is precarious -- even for the most frequent words -- we train monolingual embeddings on the Wikipedia entry for Barack Obama in English and German and compute the nearest neighbor graphs for the top 10 most frequent words, which we show in Figure \ref{fig:nn-graphs}. Even here, the graphs are not even close to being isomorphic.

%We also consider the potential impact of {\bf anchors}  in \S4.5, by simply rerunning experiments leaving out potential anchor words. 
We also investigate the sensitivity of unsupervised BDI to the {\bf dimensionality} of the monolingual word embeddings in \S\ref{sec:impact_dim}. The motivation for this is that dimensionality reduction will alter the geometric shape and remove characteristics of the embedding graphs that are important for alignment; but on the other hand, lower dimensionality introduces regularization, which will make the graphs more similar. Finally, in \S\ref{sec:impact_evaluation}, we investigate the impact of different types of query {\bf test words} on performance, including how performance varies across part-of-speech word classes and on shared vocabulary items. 

\subsection{Summary of \newcite{Lample2018crosslingual}} \label{sec:conneau_approach}
We now introduce the method of \newcite{Lample2018crosslingual}.\footnote{\url{https://github.com/facebookresearch/MUSE}} The approach builds on existing work on learning a mapping between monolingual word embeddings \cite{Mikolov2013,Xing2015} and consists of the following steps: 1) \textbf{Monolingual word embeddings:} An off-the-shelf word embedding algorithm \cite{Bojanowski2017} is used to learn source and target language spaces $X$ and $Y$. 2) \textbf{Adversarial mapping:} A translation matrix $W$ is learned between the spaces $X$ and $Y$ using adversarial techniques \cite{Ganin2016}. A discriminator is trained to discriminate samples from the translated source space $WX$ from the target space $Y$, while $W$ is trained to prevent this. This, again, is motivated by the assumption that source and target language word embeddings are approximately isomorphic. %, which we have shown not to be the case. 
3) \textbf{Refinement (Procrustes analysis):} $W$ is used to build a small bilingual dictionary of frequent words, which is pruned such that only bidirectional translations are kept \cite{IvanSeed}. A new translation matrix $W$ that translates between the spaces $X$ and $Y$ of these frequent word pairs is then induced by solving the Orthogonal Procrustes problem:

%\vspace{-1.2mm}
%\begin{footnotesize}
\begin{align}
\begin{split}
W^* = & \: \text{argmin}_W \|WX - Y\|_\text{F} = UV^\top\\
& \text{s.t.} \: U\Sigma V^\top = \text{SVD}(YX^\top)
\end{split}
\end{align}
%\end{footnotesize}

\noindent This step can be used iteratively by using the new matrix $W$ to create new seed translation pairs. It requires frequent words to serve as reliable anchors for learning a translation matrix. In the experiments in \newcite{Lample2018crosslingual}, as well as in ours, the iterative Procrustes refinement improves performance across the board. %We have also empirically verified this finding: the refinement step is \textit{always} useful, and we therefore employ it in all reported experiments.
4) \textbf{Cross-domain similarity local scaling} (CSLS) is used to expand high-density areas and condense low-density ones, for more accurate nearest neighbor calculation, CSLS reduces the hubness problem in high-dimensional spaces \cite{Radovanovic:2010jmlr,Dinu:2015arxiv}. It relies on the mean similarity of a source language embedding $x$ to its $K$ target language nearest neighbours ($K=10$ suggested) $nn_1,\ldots,nn_{K}$:

%\vspace{-1em}
%{\footnotesize
\begin{align}
mnn_{T}(x) = \frac{1}{K}\sum_{i=1}^K cos(x,nn_i)
\end{align}
where $cos$ is the cosine similarity. $mnn_{S}(y)$ is defined in an analogous manner for any target language embedding $y$. $CSLS(x,y)$ is then calculated as follows:

%\vspace{-0.7em}
%{\footnotesize
\begin{align}
2cos(x,y) - mnn_{T}(x) - mnn_{S}(y)
\end{align}

\subsection{A simple supervised method}

Instead of learning cross-lingual embeddings completely without supervision, we can extract inexpensive supervision signals by harvesting identically spelled words as in, e.g. \cite{Artetxe2017,Smith2017}. Specifically, we use identically spelled words that occur in the vocabularies of both languages as bilingual seeds, without employing any additional transliteration or lemmatization/normalization methods. Using this seed dictionary, we then run the refinement step using Procrustes analysis of \citet{Lample2018crosslingual}.

\section{Experiments} \label{sec:experiments}

%The strongest assumption that zero-supervision models make is that the distributions of the monolingual embeddings of both languages are aligned. A misalignment can be due to several factors: 1) The domains on which the monolingual embeddings were trained are different; 2) the languages of the embeddings are too dissimilar; and 3) the models used to learn the embeddings are different or use different hyperparameters.

In the following experiments, we investigate the robustness of unsupervised cross-lingual word embedding learning, varying the language pairs, monolingual corpora, hyper-parameters, etc., to obtain a better understanding of when and why unsupervised BDI works. 

\paragraph{Task: Bilingual dictionary induction}
After the shared cross-lingual space is induced, given a list of $N$ source language words $x_{u,1},\ldots,x_{u,N}$, the task is to find a target language word $t$ for each \textit{query word} $x_u$ relying on the representations in the space. $t_i$ is the target language word closest to the source language word $x_{u,i}$ in the induced cross-lingual space, also known as the {\em cross-lingual nearest neighbor}. The set of learned $N$ $(x_{u,i},t_i)$ pairs is then run against a gold standard dictionary.

We use bilingual dictionaries compiled by \newcite{Lample2018crosslingual} as gold standard, and adopt their evaluation procedure: each test set in each language consists of 1500 gold translation pairs. We rely on CSLS for retrieving the nearest neighbors, as it consistently outperformed the cosine similarity in all our experiments. Following a standard evaluation practice \cite{Vulic:2013emnlp,Mikolov2013,Lample2018crosslingual}, we report \textit{Precision at 1} scores (P@1): how many times one of the correct translations of a source word $w$ is retrieved as the nearest neighbor of $w$ in the target language.

\subsection{Experimental setup}
\label{ss:exp}
Our default experimental setup closely follows the setup of \newcite{Lample2018crosslingual}. For each language we induce monolingual word embeddings for all languages from their respective tokenized and lowercased Polyglot Wikipedias \cite{AlRfou:2013conll} using {\it fastText} \cite{Bojanowski2017}. Only words with more than 5 occurrences are retained for training. Our {\it fastText} setup relies on skip-gram with negative sampling \cite{Mikolov2013d} with standard hyper-parameters: bag-of-words contexts with the window size 2, 15 negative samples, subsampling rate $10^{-4}$, and character n-gram length 3-6. All embeddings are $300$-dimensional.

As we analyze the impact of various modeling assumptions in the following sections (e.g., domain differences, algorithm choices, hyper-parameters), we also train monolingual word embeddings using other corpora and different hyper-parameter choices. Quick summaries of each experimental setup are provided in the respective subsections.

%Finally, the list of typologically diverse test language pairs is provided in Table~\ref{langs}. A wider spectrum of typological properties covered in this work enables finer-grained analyses of unsupervised models in relation to language (dis)similarity.

\begin{table}
\begin{small}
\begin{tabular}{l|lll}
\toprule
&{\bf Marking}&{\bf Type}&{\bf \# Cases}\\
\midrule
English (\textsc{en})&dependent&isolating&None\\
French (\textsc{fr})&mixed&fusional&None\\
German (\textsc{de})&dependent&fusional&4\\
Chinese (\textsc{zh})&dependent&isolating&None\\
Russian (\textsc{ru})&dependent&fusional&6--7\\
Spanish (\textsc{es})&dependent&fusional&None\\
\midrule
Estonian (\textsc{et})&mixed&agglutinative&10+\\
Finnish (\textsc{fi})&mixed&agglutinative&10+\\
Greek (\textsc{el})&double&fusional&3\\
Hungarian (\textsc{hu})&dependent&agglutinative&10+\\
Polish (\textsc{pl})&dependent&fusional&6--7\\
Turkish (\textsc{tr})&dependent&agglutinative&6--7
\end{tabular}
\label{tab:languages}
\caption{\label{langs}Languages in \newcite{Lample2018crosslingual} and in our experiments (lower half)}
\end{small}
\end{table}

\begin{table}[t]
\centering
\def\arraystretch{1.0}
\vspace{-0.0em}
{\footnotesize
\begin{tabularx}{\linewidth}{l XXX}
\toprule
{} & {\textbf{Unsupervised}} & {\textbf{Supervised}} & {\textbf{Similarity}}\\
{} & {(Adversarial)} & {(Identical)} & {(Eigenvectors)} \\
\cmidrule(lr){2-2} \cmidrule(lr){3-3} \cmidrule(lr){4-4}
{\textsc{en-es}} & {81.89} & {\bf 82.62} &{2.07} \\
\midrule
{\textsc{en-et}} & {00.00} & {\bf 31.45} & 6.61 \\
\textsc{en-fi} & {00.09} & {\bf 28.01} & 7.33 \\
\textsc{en-el} & {00.07} & {\bf 42.96} & 5.01 \\
\textsc{en-hu} & {45.06} & {\bf 46.56} & 3.27 \\
\textsc{en-pl} & {46.83} & {\bf 52.63} & 2.56 \\
\textsc{en-tr} & {32.71} & {\bf 39.22} & 3.14 \\
\midrule
\textsc{et-fi} & {\bf 29.62} & {24.35} & 3.98 \\
\bottomrule
\end{tabularx}
}
%\vspace{-0.5mm}
\caption{Bilingual dictionary induction scores (P@1$\times$100\%) using \textbf{a)} the unsupervised method with adversarial training; \textbf{b)} the supervised method with a bilingual seed dictionary consisting of identical words (shared between the two languages). The third columns lists eigenvector similarities between 10 randomly sampled source language nearest neighbor subgraphs of 10 nodes and the subgraphs of their translations, all from the benchmark dictionaries in \newcite{Lample2018crosslingual}.}
\vspace{-0.5mm}
\label{tab:unsuper_vs_super}
\end{table}

\subsection{Impact of language similarity} \label{sec:impact_lang_sim}

\newcite{Lample2018crosslingual} present results for several target languages: Spanish, French, German, Russian, Chinese, and Esperanto. All languages but Esperanto are isolating or exclusively concatenating languages from a morphological point of view. All languages but French are dependent-marking. Table~\ref{langs} lists three important morphological properties of the languages involved in their/our experiments. 

Agglutinative languages with mixed or double marking show more morphological variance with content words, and we speculate whether unsupervised BDI is challenged by this kind of morphological complexity.  To evaluate this, we experiment with Estonian and Finnish, and we include Greek, Hungarian, Polish, and Turkish to see how their approach fares on combinations of these two morphological traits. 

%The promise of zero-supervision models is that we can learn cross-lingual embeddings even for low-resource languages. On the other hand, a similar distribution of embeddings requires languages to be similar.

%To test to what extent the quality of the induced embeddings depend on the similarity of the languages, we learn monolingual embeddings of the members of the West-Germanic (English), Romance (Spanish), Finnic (Finnish), and Turkic (Turkish) language families on Wikipedia. We induce cross-lingual spaces between each language pair and evaluate the spaces on BLI. Our assumption is that embedding spaces are more reliable between similar languages, i.e. English and Spanish as members of the Indo-European language family.

We show results in the left column of Table~\ref{tab:unsuper_vs_super}.  The results are quite dramatic. The approach achieves impressive performance for Spanish, one of the languages \newcite{Lample2018crosslingual} include in their paper. For the languages we add here, performance is less impressive. For the languages with dependent marking (Hungarian, Polish, and Turkish), P@1 scores are still reasonable, with Turkish being slightly lower (0.327) than the others. However, for Estonian and Finnish, the method fails completely. Only in less than 1/1000 cases does a nearest neighbor search in the induced embeddings return a correct translation of a query word.\footnote{We note, though, that varying our random seed, performance for Estonian, Finnish, and Greek is sometimes (approximately 1 out of 10 runs) {\em on par}~with Turkish. Detecting main causes and remedies for the inherent instability of adversarial training is one the most important avenues for future research.}

The sizes of Wikipedias naturally vary across languages: e.g., {\it fastText} trains on approximately 16M sentences and 363M word tokens for Spanish, while it trains on 1M sentences and 12M words for Finnish. However, the difference in performance cannot be explained by the difference in training data sizes. To verify that near-zero performance in Finnish is not a result of insufficient training data, we have conducted another experiment using the large {\bf Finnish WaC corpus} \cite{Ljubesic:2016fiwac} containing 1.7B words in total (this is similar in size to the English Polyglot Wikipedia). However, even with this large Finnish corpus, the model does not induce anything useful: P@1 equals 0.0.

We note that while languages with mixed marking may be harder to align, it seems unsupervised BDI is possible between similar, mixed marking languages. So while unsupervised learning fails for English-Finnish and English-Estonian, performance is reasonable and stable for the more similar Estonian-Finnish pair (Table~\ref{tab:unsuper_vs_super}). In general, unsupervised BDI, using the approach in \newcite{Lample2018crosslingual}, seems challenged when pairing English with languages that are not isolating and do not have dependent marking.\footnote{One exception here is French, which they include in their paper, but French arguably has a relatively simple morphology.}

The promise of zero-supervision models is that we can learn cross-lingual embeddings even for low-resource languages. On the other hand, a similar distribution of embeddings requires languages to be similar. This raises the question whether we need fully unsupervised methods at all. In fact, our supervised method that relies on very naive supervision in the form of identically spelled words leads to competitive performance for similar language pairs and better results for dissimilar pairs. The fact that we can reach competitive and more robust performance with such a simple heuristic questions the true applicability of fully unsupervised approaches and suggests that it might often be better to rely on available weak supervision.

%While it is reasonable to assume that such unsupervised cross-lingual  methods will be more effective for similar language pairs, an important question is whether we need fully unsupervised methods for such languages at all. Inexpensive supervision signals may be extracted for such languages by just harvesting identically spelled words as done in, e.g. \cite{Artetxe2017,Smith2017}. As indicated by the results in Table~\ref{tab:unsuper_vs_super}, this leads to competitive performance for similar language pairs, and better results for dissimilar pairs, questioning the true applicability of fully unsupervised approaches.

\captionsetup[subfigure]{oneside,margin={-0.1cm,0.6cm},skip=5.5pt}
\begin{figure*}[!t]
    \centering
    \begin{subfigure}[t]{0.32\linewidth}
        \centering
        \includegraphics[width=0.96\linewidth]{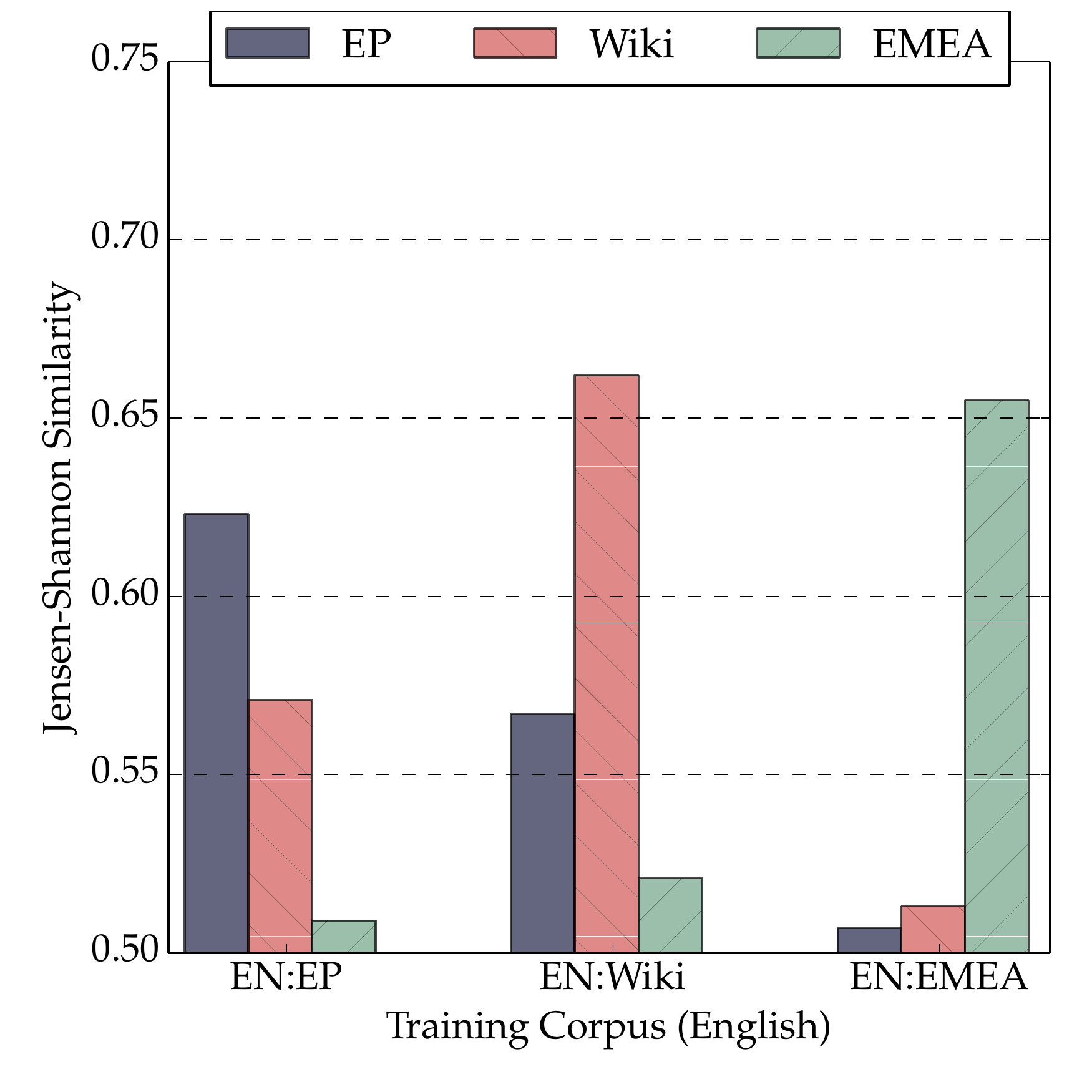}
        \vspace{-0.3em}
        \caption{en-es: \textit{domain similarity}}
        \label{fig:glove}
    \end{subfigure}
    \begin{subfigure}[t]{0.32\textwidth}
        \centering
        \includegraphics[width=0.96\linewidth]{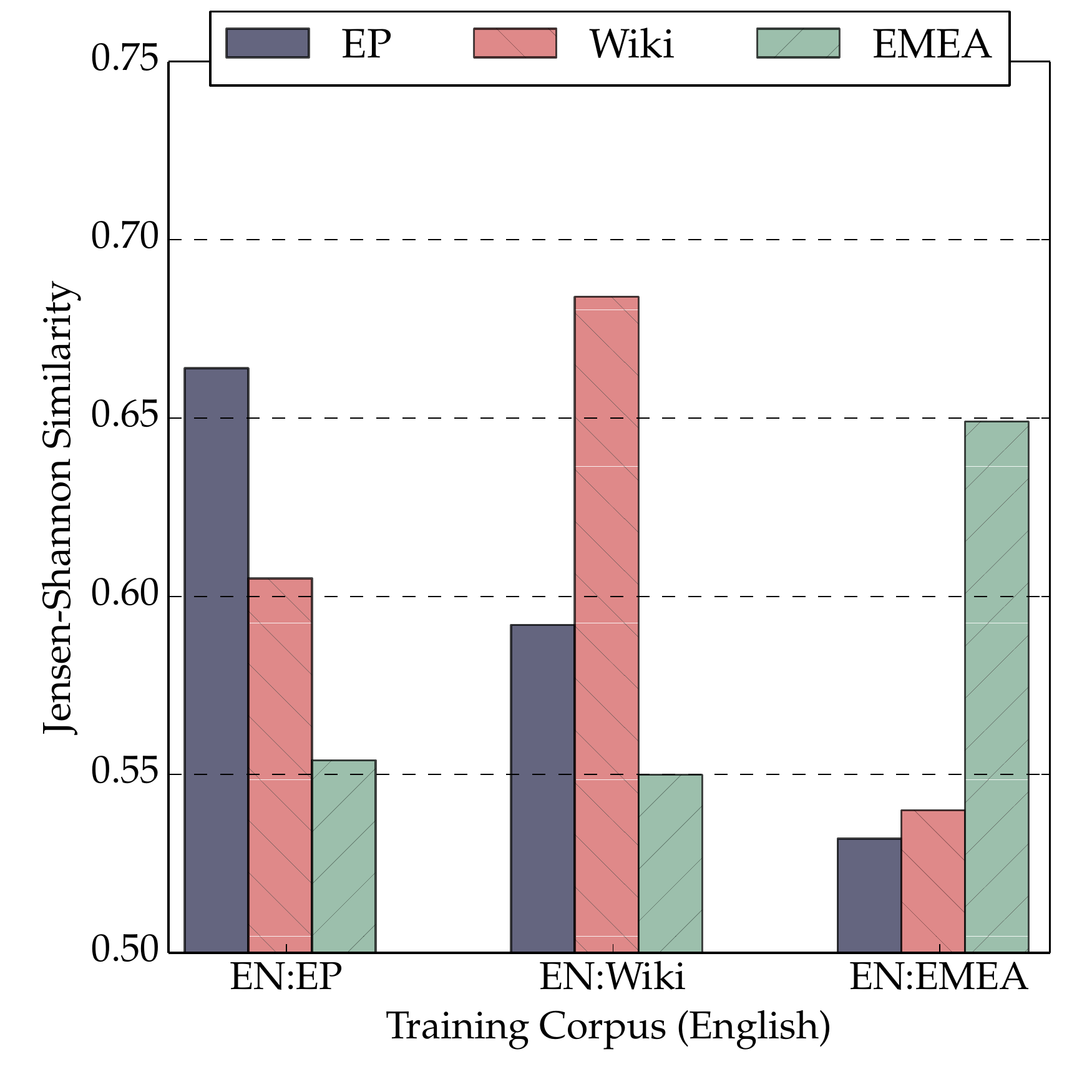}
        \vspace{-0.3em}
        \caption{en-fi: \textit{domain similarity}}
        \label{fig:bow2}
    \end{subfigure}
    \begin{subfigure}[t]{0.32\textwidth}
        \centering
        \includegraphics[width=0.96\linewidth]{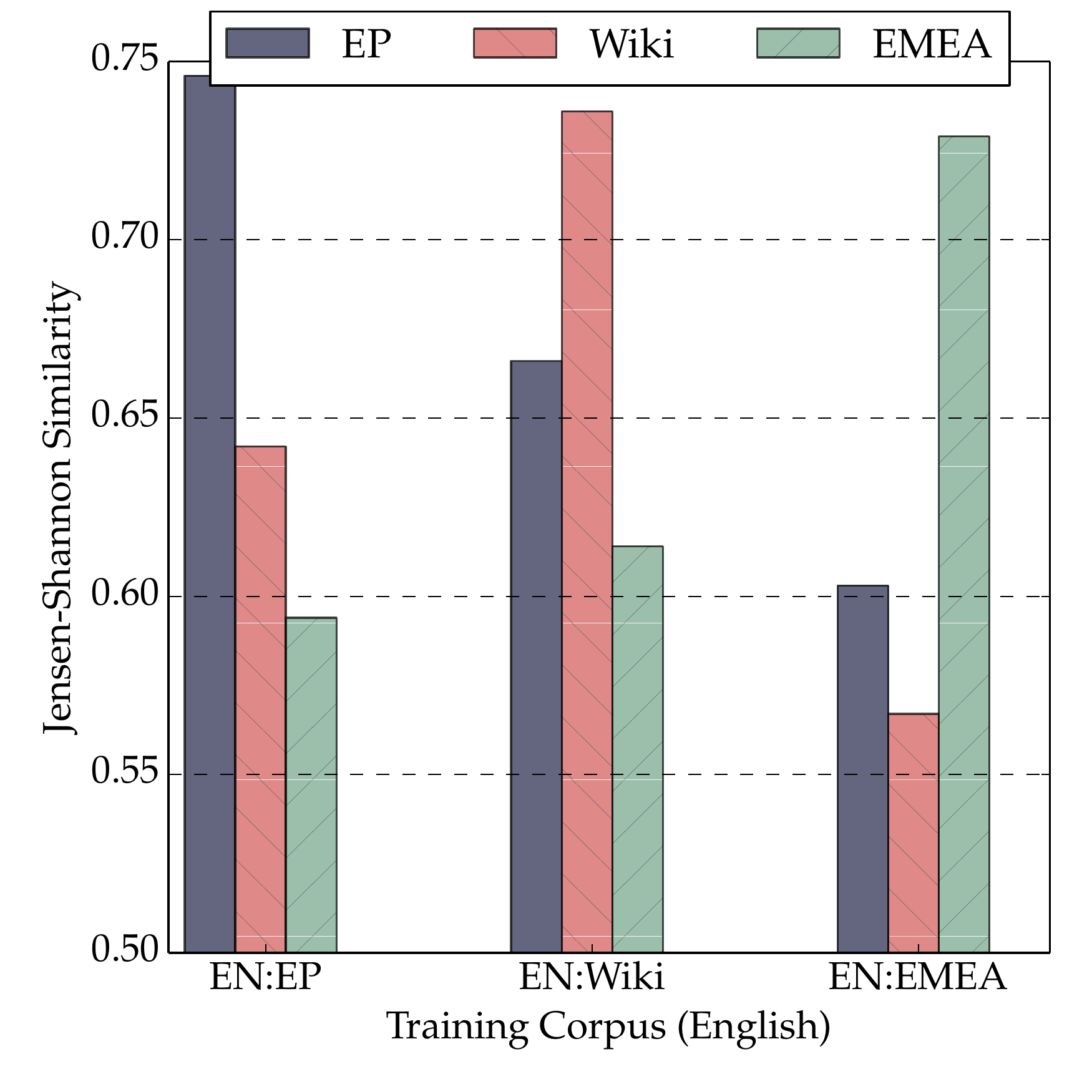}
        \vspace{-0.3em}
        \caption{en-hu: \textit{domain similarity}}
        \label{fig:ft}
    \end{subfigure}
    \begin{subfigure}[t]{0.32\linewidth}
        \centering
        \includegraphics[width=0.96\linewidth]{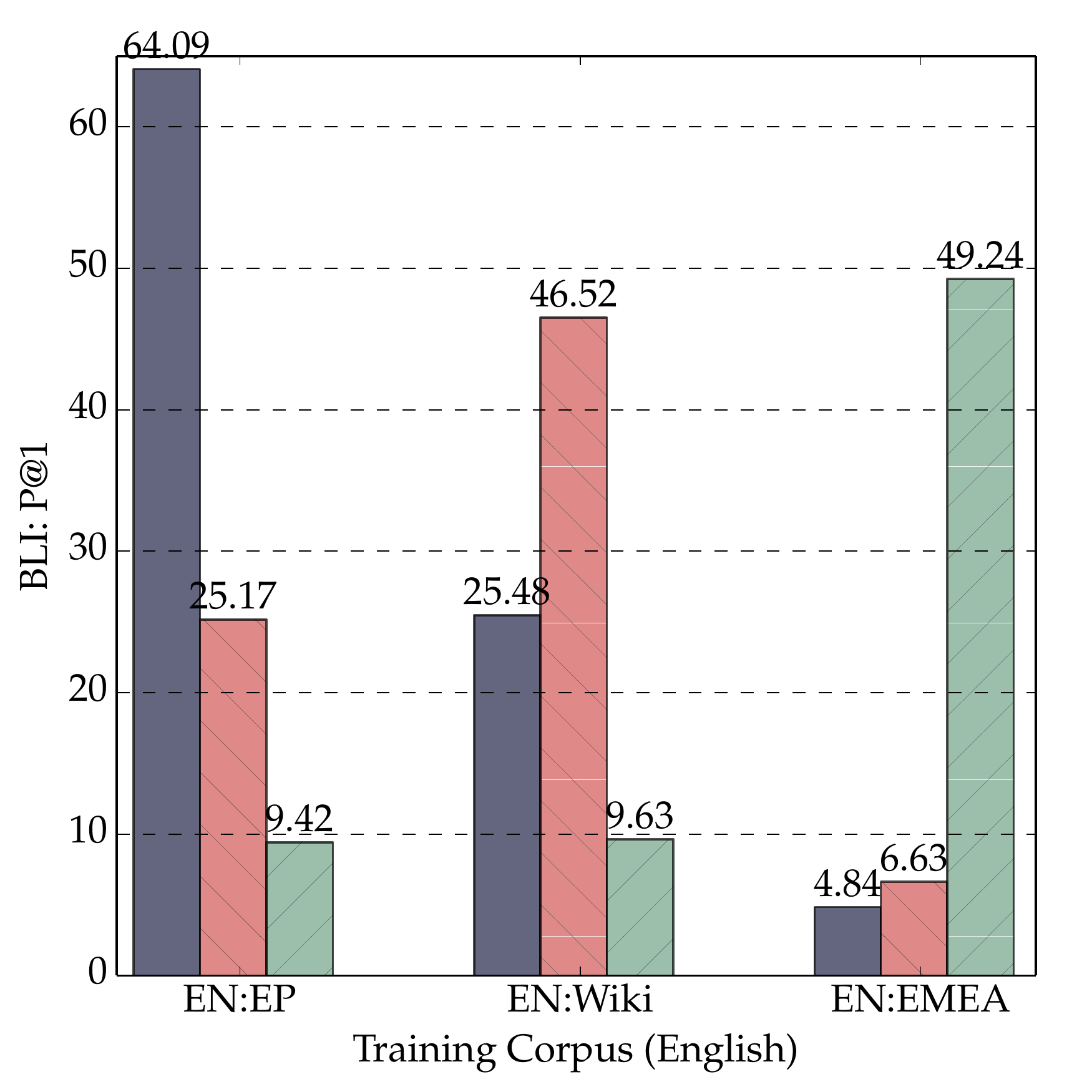}
        \vspace{-0.3em}
        \caption{en-es: \textit{identical words}}
        \label{fig:glove}
    \end{subfigure}
    \begin{subfigure}[t]{0.32\textwidth}
        \centering
        \includegraphics[width=0.96\linewidth]{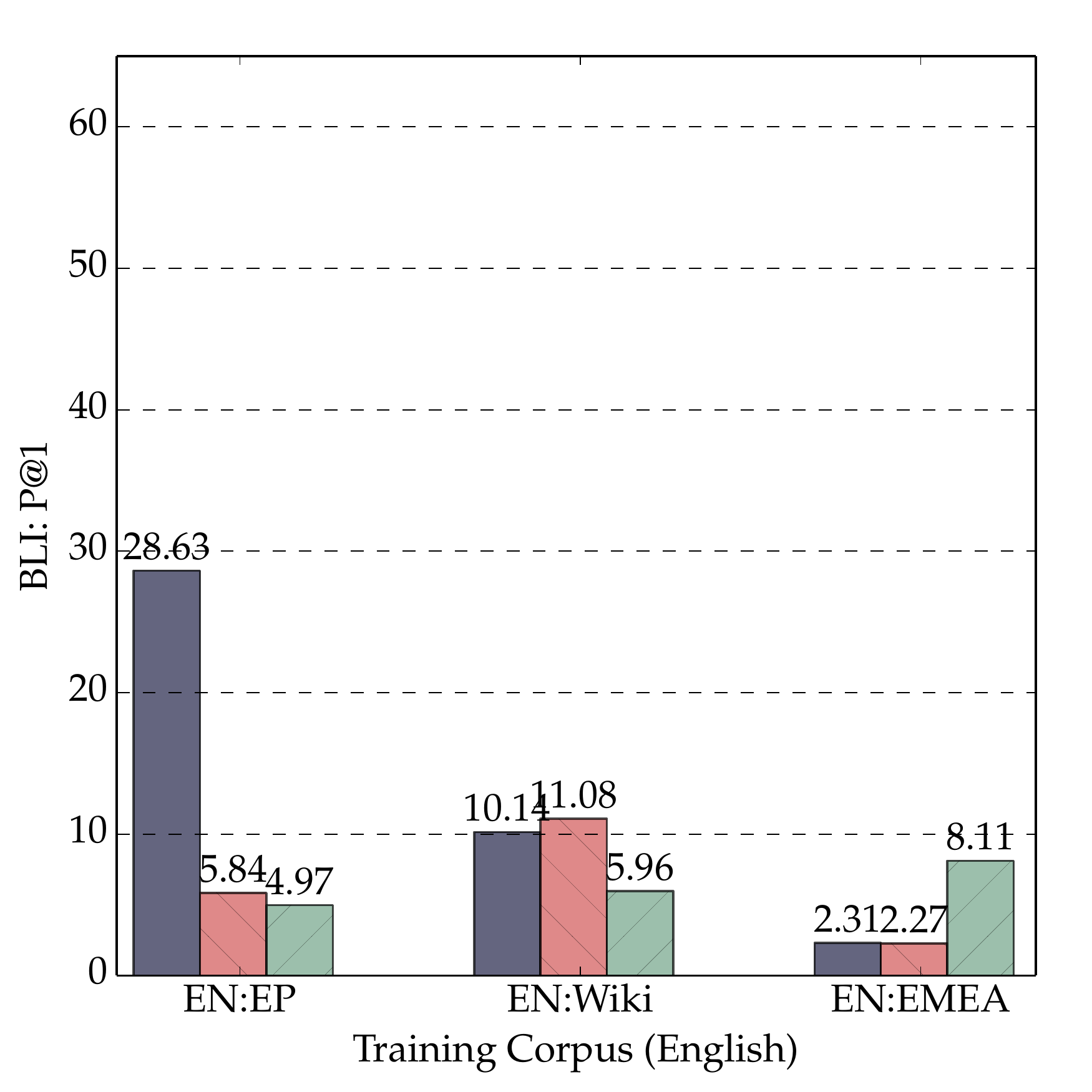}
        \vspace{-0.3em}
        \caption{en-fi: \textit{identical words}}
        \label{fig:bow2}
    \end{subfigure}
    \begin{subfigure}[t]{0.32\textwidth}
        \centering
        \includegraphics[width=0.96\linewidth]{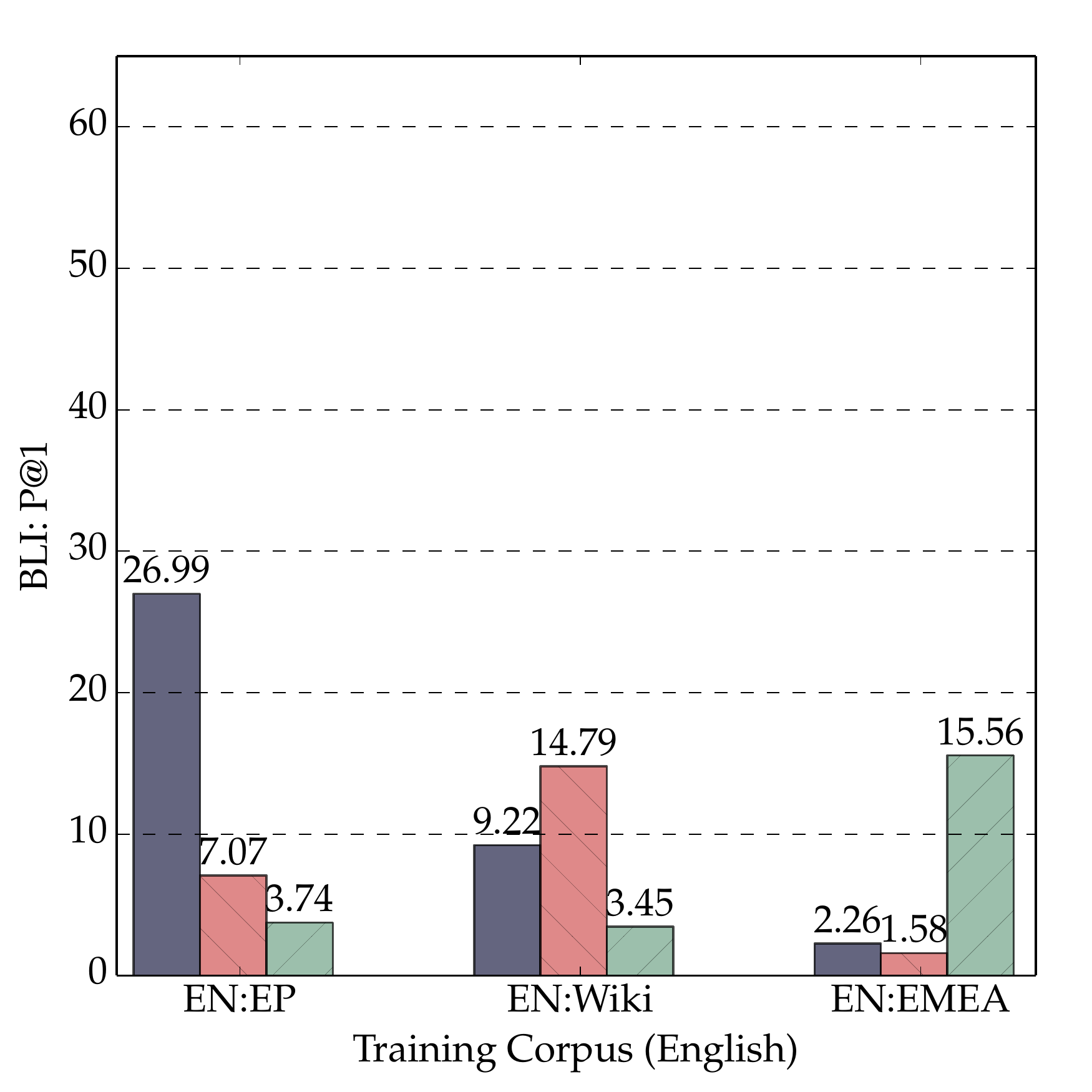}
        \vspace{-0.3em}
        \caption{en-hu: \textit{identical words}}
        \label{fig:ft}
    \end{subfigure}
    \begin{subfigure}[t]{0.32\linewidth}
        \centering
        \includegraphics[width=0.96\linewidth]{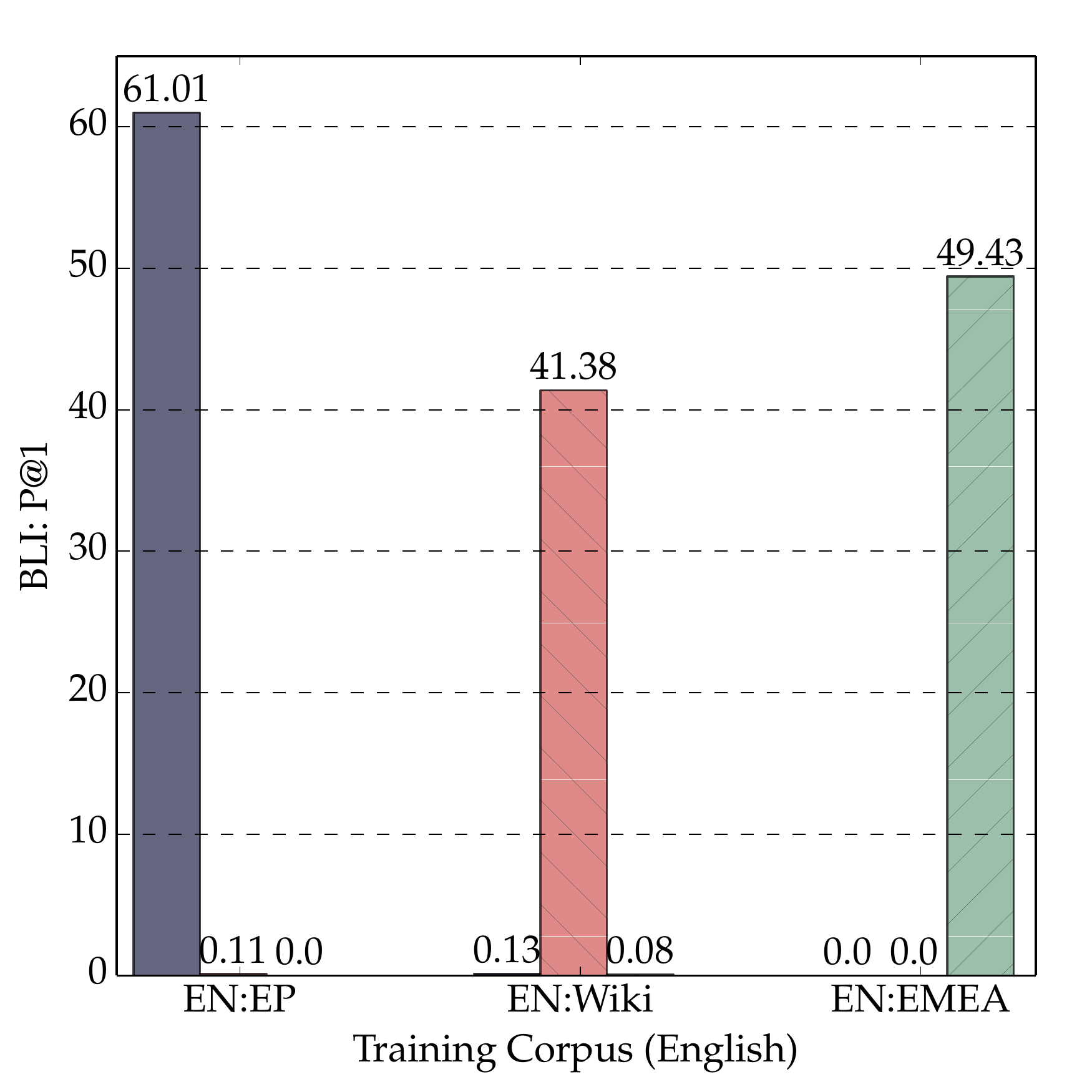}
        \vspace{-0.3em}
        \caption{en-es: \textit{fully unsupervised BLI}}
        \label{fig:glove}
    \end{subfigure}
    \begin{subfigure}[t]{0.32\textwidth}
        \centering
        \includegraphics[width=0.96\linewidth]{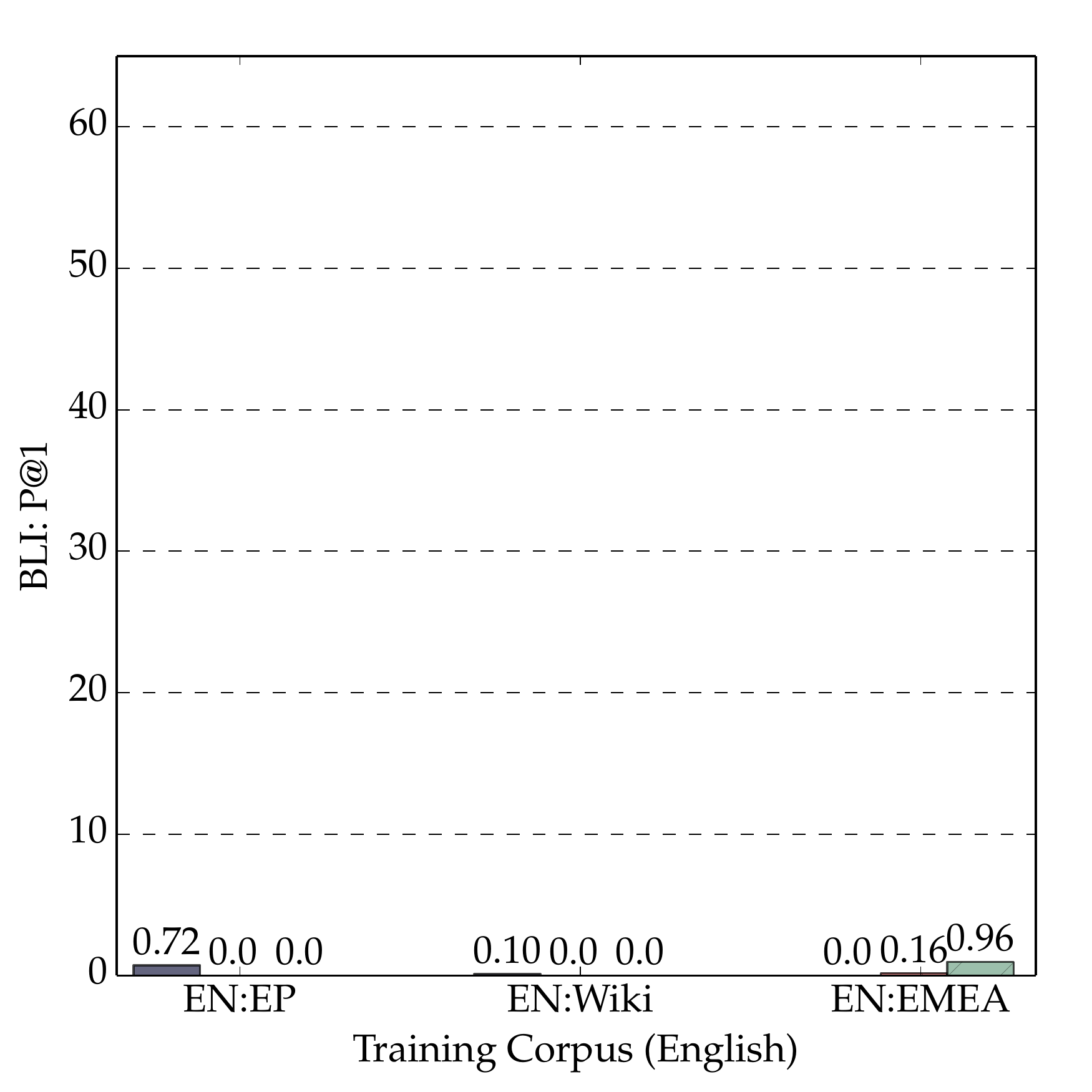}
        \vspace{-0.3em}
        \caption{en-fi: \textit{fully unsupervised BLI}}
        \label{fig:bow2}
    \end{subfigure}
    \begin{subfigure}[t]{0.32\textwidth}
        \centering
        \includegraphics[width=0.96\linewidth]{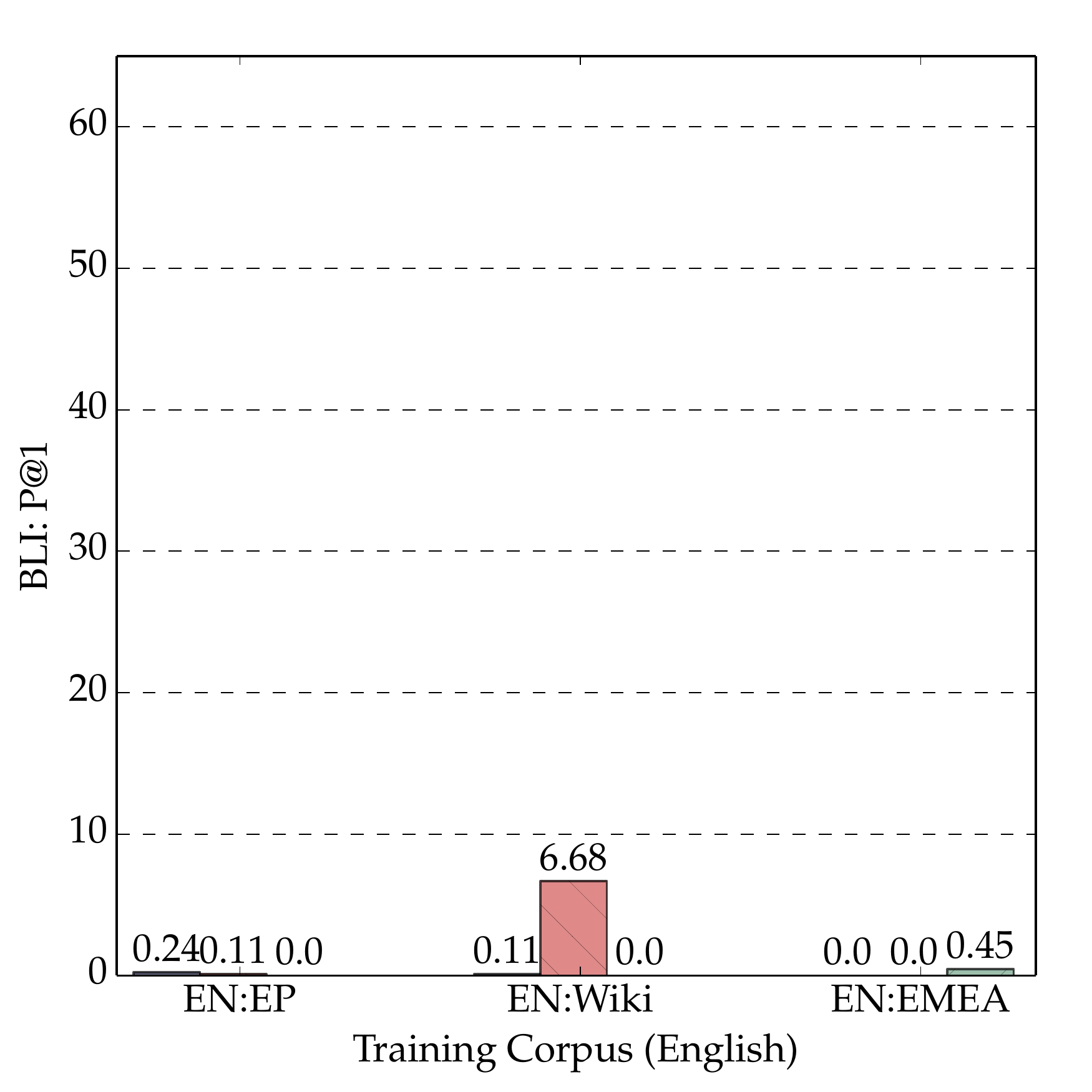}
        \vspace{-0.5em}
        \caption{en-hu: \textit{fully unsupervised BLI}}
        \label{fig:ft}
    \end{subfigure}
    %\vspace{-1.5mm}
    \caption{Influence of language-pair \textit{and} domain similarity on BLI performance, with three language pairs (en-es/fi/hu). \textbf{Top row, (a)-(c)}: Domain similarity (higher is more similar) computed as $dsim=1-JS$, where $JS$ is Jensen-Shannon divergence; \textbf{Middle row, (d)-(f)}: baseline BLI model which learns a linear mapping between two monolingual spaces based on a set of identical (i.e., shared) words; \textbf{Bottom row, (g)-(i)}: fully unsupervised BLI model relying on the distribution-level alignment and adversarial training. Both BLI models apply the Procrustes analysis and use CSLS to retrieve nearest neighbours.}
%\vspace{-0.5mm}
\label{fig:domain}
\end{figure*}

\subsection{Impact of domain differences} \label{sec:impact_domain_diff}

Monolingual word embeddings used in \newcite{Lample2018crosslingual} are induced from Wikipedia, a near-parallel corpus. In order to assess the sensitivity of unsupervised BDI to the comparability and domain similarity of the monolingual corpora, we replicate the experiments in \newcite{Lample2018crosslingual} using combinations of word embeddings extracted from three different domains: \textbf{1)} parliamentary proceedings from EuroParl.v7 \cite{Koehn:2005}, \textbf{2)} Wikipedia \cite{AlRfou:2013conll}, and \textbf{3)} the EMEA corpus in the medical domain \cite{Tiedemann:2009opus}. We report experiments with three language pairs: English-$\{$Spanish, Finnish, Hungarian$\}$.

To control for the corpus size, we restrict each corpus in each language to 1.1M sentences in total (i.e., the number of sentences in the smallest, EMEA corpus). $300$-dim {\it fastText} vectors are induced as in \S\ref{ss:exp}, retaining all words with more than 5 occurrences in the training data. For each pair of monolingual corpora, we compute their domain (dis)similarity by calculating the Jensen-Shannon divergence \cite{El-Gamal1991}, based on term distributions.\footnote{In order to get comparable term distributions, we translate the source language to the target language using the bilingual dictionaries provided by \citeauthor{Lample2018crosslingual} \shortcite{Lample2018crosslingual}.} The domain similarities are displayed in Figures~\ref{fig:domain}a--c.\footnote{We also computed $\mathcal{A}$-distances \cite{Blitzer2007} and confirmed that trends were similar.}

%\paragraph{Different domains} In order to assess whether zero-supervision models are able to deal with a domain shift, we learn monolingual embeddings based on three different domains: Wikipedia, Europarl, and the Bible. We then learn a cross-lingual embedding space between each domain pair, yielding 9 spaces and evaluate each space using BLI. 
We show the results of unsupervised BDI in Figures~\ref{fig:domain}g--i. For Spanish, we see good performance in all three cases where the English and Spanish corpora are from the same domain. {\em When the two corpora are from different domains, performance is close to zero.} For Finnish and Hungarian, performance is always poor, suggesting that more data is needed, even when domains are similar. This is in sharp contrast with the results of our minimally supervised approach (Figures~\ref{fig:domain}d--f) based on identical words, which achieves decent performance in many set-ups.

We also observe a strong decrease in P@1 for English-Spanish (from 81.19\% to 46.52\%) when using the smaller Wikipedia corpora. This result indicates the importance of procuring large monolingual corpora from similar domains in order to enable unsupervised dictionary induction. However, resource-lean languages, for which the unsupervised method was designed in the first place, cannot be guaranteed to have as large monolingual training corpora as available for English, Spanish or other major resource-rich languages.

%\begin{table}[]
%\centering
%\begin{tabular}{l c c c}
%\toprule
%src $\downarrow$ \: trg $\rightarrow$ & Wikipedia & Europarl & Bible \\
%\midrule
%Wikipedia &  &  &  \\
%Europarl &  &  &  \\
%Bible &  &  & \\
%\bottomrule
%\end{tabular}
%\caption{BLI results for cross-lingual embeddings trained on different domains.}
%\label{tab:different-domains}
%\end{table}

\subsection{Impact of hyper-parameters} \label{sec:impact_hyperparam} % algorithm choice is also a hyper-parameter

%Another source of misalignment between distributions may be the architecture and the hyperparameters used to learn the monolingual embedding models. We first train monolingual embeddings on Wikipedia using fastText \cite{Bojanowski2017}, word2vec \cite{Mikolov2013d}, and GloVe \cite{Pennington2014}, induce cross-lingual spaces, and evaluate on BLI. We show results in Table~\ref{tab:different-models}.

\newcite{Lample2018crosslingual} use the same hyper-parameters for inducing embeddings for all languages. This is of course always practically possible, but we are interested in seeing whether their approach works on pre-trained embeddings induced with possibly very different hyper-parameters. We focus on two hyper-parameters: context window-size (\textit{win}) and the parameter controlling the number of $n$-gram features in the {\it fastText} model (\textit{chn}), while at the same time varying the underlying algorithm: \textit{skip-gram} vs. \textit{cbow}. The results for English-Spanish are listed in Table~\ref{tab:impact_hyper}. 

The small variations in the hyper-parameters with the same underlying algorithm (i.e., using \textit{skip-gram} or \textit{cbow} for both \textsc{en} and \textsc{es}) yield only slight drops in the final scores. Still, the best scores are obtained with the same configuration on both sides. Our main finding here is that unsupervised BDI fails (even) for \textsc{en-es} when the two monolingual embedding spaces are induced by two different algorithms (see the results of the entire Spanish \textit{cbow} column).\footnote{We also checked if this result might be due to a lower-quality monolingual \textsc{es} space. However, monolingual word similarity scores on available datasets in Spanish show performance comparable to that of Spanish \textit{skip-gram} vectors: e.g., Spearman's $\rho$ correlation is $\approx 0.7$ on the \textsc{es} evaluation set from SemEval-2017 Task 2 \cite{Camacho:2017semeval}.} In sum, this means that {\em the unsupervised approach is unlikely to work on pre-trained word embeddings unless they are induced on same- or comparable-domain, reasonably-sized training data using the same underlying algorithm}.

\begin{table}[t]
\centering
\def\arraystretch{1.0}
\vspace{-0.0em}
{\footnotesize
\begin{tabularx}{\linewidth}{l XX}
\toprule
{} & \multicolumn{2}{c}{\textbf{English}} \\
{} & \multicolumn{2}{c}{(skipgram, win=2, chn=3-6)} \\
\cmidrule(lr){2-3}
{} & {\textbf{Spanish}} & {\textbf{Spanish}} \\
{} & {\it (skipgram)} & {\it (cbow)} \\
\cmidrule(lr){2-2} \cmidrule(lr){3-3}
{==} & {81.89} & {00.00} \\
{$\neq$ win=10} & {81.28} & {00.07} \\
{$\neq$ chn=2-7} & {80.74} & {00.00} \\
{$\neq$ win=10, chn=2-7} & {80.15} & {00.13} \\

%\textbf{Random-Init} of $\mathcal{V}_d$ & {.809} & {.809} \\
%\midrule
\bottomrule
\end{tabularx}
}
%\vspace{-0.5em}
\caption{Varying the underlying {\it fastText} algorithm and hyper-parameters. The first column lists differences in training configurations between English and Spanish monolingual embeddings.}
%\vspace{-1.5mm}
\label{tab:impact_hyper}
\end{table}

%\subsection{Impact of anchors} \label{sec:anchors}

%For the language pairs, where \newcite{Lample2018crosslingual} achieve good or reasonable performance, it is interesting to see whether removing potential anchors such as punctuation and dummy symbols leads to worse representation and lower performance on bilingual dictionary induction. We performed such experiments for a representative set of languages (Spanish, Hungarian, and Finnish), but did not see changes in performance. {\bf [UPDATE WITH ACTUAL RESULTS?]}

\subsection{Impact of dimensionality} \label{sec:impact_dim}
We also perform an experiment on $40$-dimensional monolingual word embeddings. This leads to reduced expressivity, and can potentially make the geometric shapes of embedding spaces harder to align; on the other hand, reduced dimensionality may also lead to less overfitting. We generally see worse performance (P@1 is 50.33 for Spanish, 21.81 for Hungarian, 20.11 for Polish, and 22.03 for Turkish) -- but, very {\em interestingly}, we obtain {\em better performance} for Estonian (13.53), Finnish (15.33), and Greek (24.17) than we did with 300 dimensions. We hypothesize this indicates monolingual word embedding algorithms over-fit to some of the rarer peculiarities of these languages.  
%on PW, P@1: EN-ES: 50.33; EN-FI: 15.33, EN-HU: 21.81, EN-EL: 24.17, EN-ET: 13.53, EN-TR: 22.03, EN-PL: 20.11

\subsection{Impact of evaluation procedure} \label{sec:impact_evaluation}

BDI models are evaluated on a held-out set of query words. Here, we analyze the performance of the unsupervised approach across different parts-of-speech, frequency bins, and with respect to query words that have orthographically identical counterparts in the target language with the same or a different meaning.  

%\vspace{1.4mm}
\paragraph{Part-of-speech} We show the impact of the part-of-speech of the query words in Table~\ref{tab:query-pos}; again on a representative subset of our languages. The results indicate that performance on verbs is lowest across the board. This is consistent with research on distributional semantics and verb meaning \cite{Schwartz:2015conll,Gerz:2016emnlp}. 

\begin{table}[]
\centering
{\footnotesize
\def\arraystretch{0.93}
\begin{tabularx}{\linewidth}{l XXX}
\toprule
 & en-es & en-hu & en-fi \\
\cmidrule(lr){2-2} \cmidrule(lr){3-3} \cmidrule(lr){4-4}
Noun & 80.94 & 26.87 & 00.00\\
Verb & 66.05 & 25.44 & 00.00 \\
Adjective & 85.53 & 53.28 & 00.00\\
Adverb & 80.00 & 51.57 & 00.00\\
Other & 73.00 & 53.40 & 00.00\\
\bottomrule
\end{tabularx}
}%
\caption{P@$1\times 100\%$ scores for query words with different parts-of-speech.}
\label{tab:query-pos}
\end{table}

\begin{figure}[t!]
        \centering
       \includegraphics[width=0.75\linewidth]{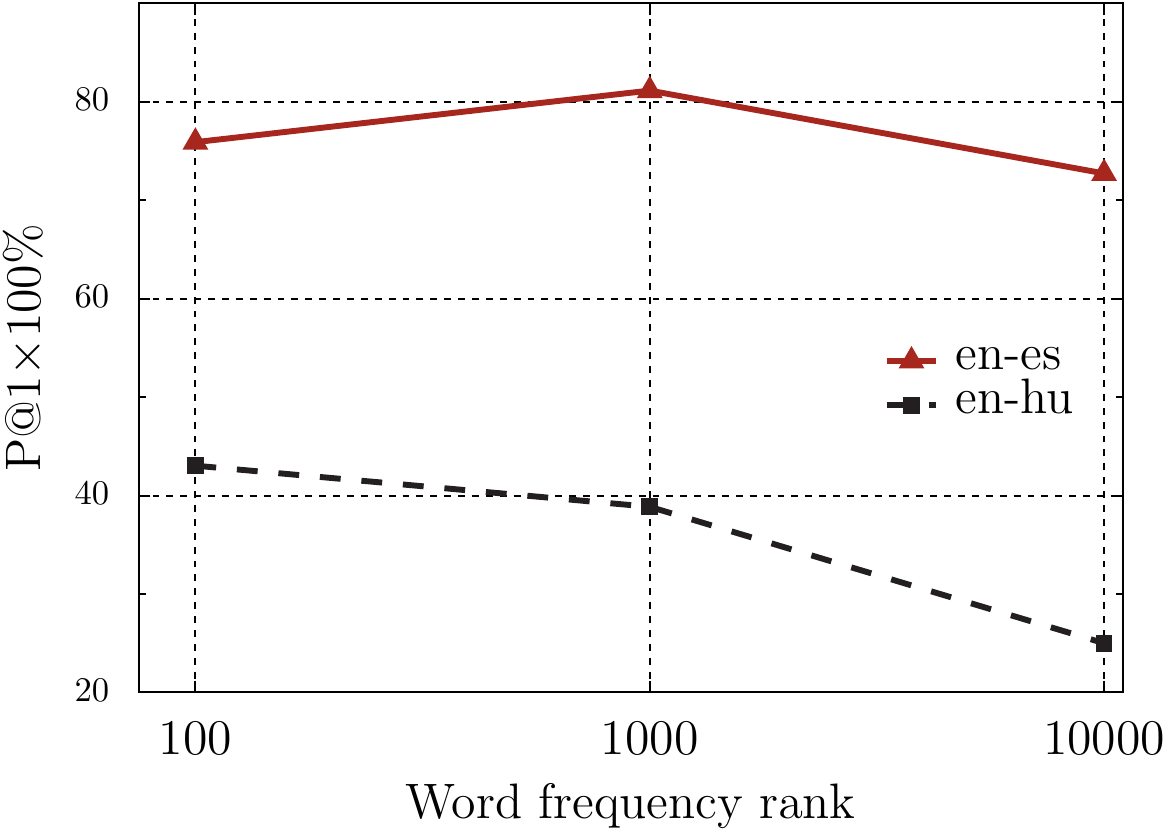}
         \caption{P@1 scores for \textsc{en-es} and \textsc{en-hu} for queries with different frequency ranks.}
         \label{fig:freq_plot}
         \vspace{-2mm}
\end{figure}

%\vspace{1.4mm}
\paragraph{Frequency} We also investigate the impact of the frequency of query words. We calculate the word frequency of English words based on Google's Trillion Word Corpus: query words are divided in groups based on their rank -- i.e., the first group contains the top 100 most frequent words, the second one contains the 101th-1000th most frequent words, etc. -- and plot performance (P@1) relative to rank in Figure~\ref{fig:freq_plot}. For \textsc{en-fi}, P@1 was 0 across all frequency ranks. The plot shows sensitivity to frequency for \textsc{hu}, but less so for \textsc{es}.

\paragraph{Homographs} Since we use identical word forms (homographs) for supervision, we investigated whether these are representative or harder to align than other words. Table~\ref{tab:query-homograph} lists performance for three sets of query words: (a) source words that have homographs (words that are spelled the same way) with the same meaning (homonyms) in the target language, e.g., many proper names; (b) source words that have homographs that are not homonyms in the target language, e.g., many short words; and (c) other words. Somewhat surprisingly, words which have translations that are homographs, are associated with {\em lower}~precision than other words. This is probably due to loan words and proper names, but note that using homographs as supervision for alignment, we achieve high precision for this part of the vocabulary {\em for free}. 

\begin{table}[]
\centering
{\footnotesize
\def\arraystretch{0.93}
\begin{tabularx}{\linewidth}{X X X X X}
\toprule
Spelling & Meaning & en-es & en-hu & en-fi \\
\midrule
Same & Same & 45.94 & 18.07 & 00.00\\
Same & Diff & 39.66 & 29.97 & 00.00 \\
Diff & Diff & 62.42 & 34.45 & 00.00\\
\bottomrule
\end{tabularx}}%
\caption{Scores (P@$1\times 100\%$) for query words with same and different spellings and meanings.}
\label{tab:query-homograph}
\vspace{-2.5mm}
\end{table}

\subsection{Evaluating eigenvector similarity} \label{sec:evaluation_eigenvector}

\begin{figure}
\centering
\includegraphics[width=0.8\linewidth]{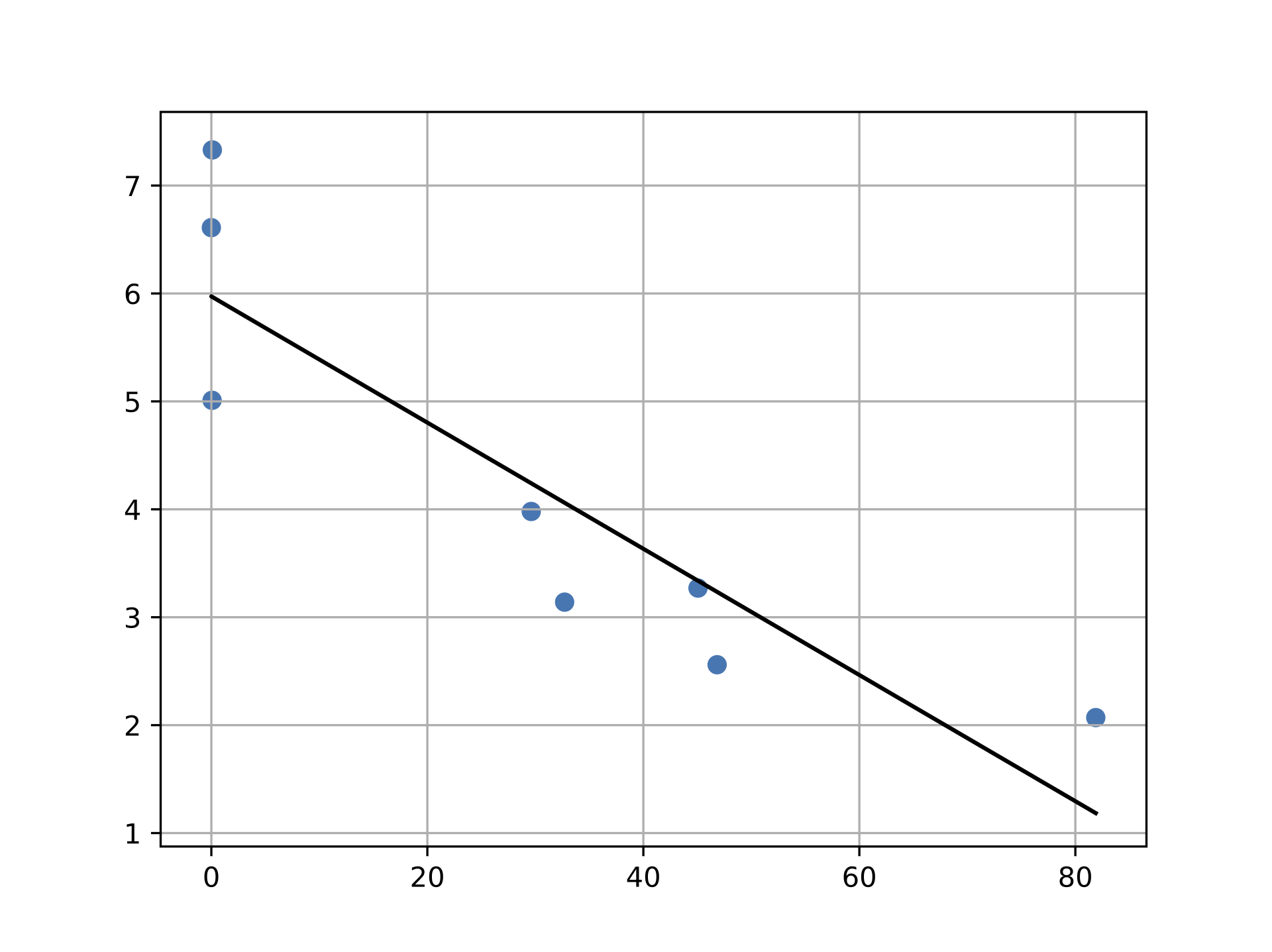}
\vspace{-0.7em}
\caption{\label{cor}Strong correlation ($\rho=0.89$) between BDI performance ($x$) and graph similarity ($y$)}
\vspace{-1.5mm}
\end{figure}

Finally, in order to get a better understanding of the limitations of unsupervised BDI, we correlate the graph similarity metric described in \S\ref{sec:how_similar} (right column of Table~\ref{tab:unsuper_vs_super}) with performance across languages (left column). Since we already established that the monolingual word embeddings are far from isomorphic---in contrast with the intuitions motivating previous work \cite{Mikolov2013,Barone2016,Zhang2017c,Lample2018crosslingual}---we would like to establish another diagnostic metric that identifies embedding spaces for which the approach in \newcite{Lample2018crosslingual} is likely to work. Differences in morphology, domain, or embedding parameters seem to be predictive of poor performance, but a metric that is independent of linguistic categorizations and the characteristics of the monolingual corpora would be more widely applicable. We plot the values in Table~\ref{tab:unsuper_vs_super} in Figure~\ref{cor}. Recall that our graph similarity metric returns a value in the half-open interval $[0,\infty)$. The correlation between BDI performance and graph similarity is strong ($\rho\sim 0.89$).

\section{Related work} \label{sec:related_work}

\paragraph{Cross-lingual word embeddings} Cross-lingual word embedding models typically, unlike \newcite{Lample2018crosslingual}, require aligned words, sentences, or documents \cite{Levy:ea:17}. Most approaches based on word alignments learn an explicit mapping between the two embedding spaces \cite{Mikolov2013,Xing2015}. %Most word-aligned models learn a mapping \cite{Mikolov2013} between source and target embeddings with an orthogonality constraint on the induced translation matrix \cite{Xing2015}, while sentence-aligned models extend successful monolingual models to the bilingual setting 
%\cite{Chandar2014,Gouws2015}. 
Recent approaches try to minimize the amount of supervision needed \cite{IvanSeed,Artetxe2017,Smith2017}. See \newcite{Upadhyay:2016acl} and \newcite{Ruder2018survey} for surveys.

\paragraph{Unsupervised cross-lingual learning} \newcite{Haghighi2008} were first to explore unsupervised BDI, using features such as context counts and orthographic substrings, and canonical correlation analysis. Recent approaches use adversarial learning \cite{goodfellow2014generative} and employ a discriminator, trained to distinguish between the translated source and the target language space, and a generator learning a translation matrix \cite{Barone2016}. \newcite{Zhang2017c}, in addition, use different forms of regularization for convergence, while \newcite{Lample2018crosslingual} uses additional steps to refine the induced embedding space.

\paragraph{Unsupervised machine translation} Research on unsupervised machine translation \cite{Lample2017,ArtetxeNMT,Lample:2018new} has generated a lot of interest recently with a promise to support the construction of MT systems for and between resource-poor languages. All unsupervised NMT methods critically rely on accurate unsupervised BDI and back-translation. Models are trained to reconstruct a corrupted version of the source sentence and to translate its translated version back to the source language. Since the crucial input to these systems are indeed cross-lingual word embedding spaces induced in an unsupervised fashion, in this paper we also implicitly investigate one core limitation of such unsupervised MT techniques. %\newcite{Lample2017} augment this set-up with an adversarial training objective.

\section{Conclusion}

We investigated when unsupervised BDI~\cite{Lample2018crosslingual} is possible and found that differences in morphology, domains or word embedding algorithms may challenge this approach. Further, we found eigenvector similarity of sampled nearest neighbor subgraphs to be predictive of unsupervised BDI performance. We hope that this work will guide further developments in this new and exciting field.

\section*{Acknowledgments}

We thank the anonymous reviewers, as well as Hinrich Sch\"{u}tze and Yova Kementchedjhieva, for their valuable feedback. Anders is supported by the ERC Starting Grant LOWLANDS No.~313695 and a Google Focused Research Award. Sebastian is supported by Irish Research Council Grant Number EBPPG/2014/30 and Science Foundation Ireland Grant Number SFI/12/RC/2289. Ivan is supported by the ERC Consolidator Grant LEXICAL No.~648909.

% include your own bib file like this:
%\bibliographystyle{acl}
%\bibliography{naaclhlt2018}
\bibliography{unsupervised_xlingual}
\bibliographystyle{acl_natbib}

\end{document}